# FOON Creation and Traversal for Recipe Generation

Raj Patel
College of Engineering
University of South Florida
Tampa, United States

*Abstract*—Task competition by robots is still off from being completely dependable and useable. One way a robot may decipher information given to it and accomplish tasks is by utilizing FOON, which stands for *functional object-oriented network*. The network first needs to be created by having a human creates action nodes as well as input and output nodes in a .txt file. After the network is sizeable, utilization of this network allows for traversal of the network in a variety of ways such as choosing steps via iterative deepening searching by using the first seen valid option. Another mechanism is heuristics, such as choosing steps based on the highest success rate or lowest amount of input ingredients. Via any of these methods, a program can traverse the network given an output product, and derive the series of steps that need to be taken to produce the output.

## I. INTRODUCTION

The motivation behind utilizing the FOON network for robotic recipe creation (or any other step related taskwork that can be derived or implemented with this network) involves the notion that robots are becoming more prevalent in the lives of humans, but a problem remains of how is there a way a robot can decide for itself the steps it needs to do to perform a task without direct human input? Of course, given a series of steps, a robot will execute them as it is its sole purpose. However, this situation requires a human to provide the steps that the robot can do when it would be better if the robot itself were "smart" in that it would already have the data to compute what steps it has to do to get the job done. Such an improvement would allow humans to be relatively hands-off and prevent manual reconfiguration of the robot's instructions for every single new task to be done. FOON serves as a potential solution to this problem as it provides a network that the robot or program can traverse to get the series of steps needed to produce the final end product [1]. In the experiment discussed in this report, the main task of the robot is to create a dish by determining what steps need to be taken to produce that dish. FOON aids in providing a network that the robot can traverse given a variety of traversal methods in order to reach the steps its need to make the disk.

## II. TERMINOLOGY

### A. Object Node

The object node is one of the two fundamental units within FOON. An object node is exactly what it sounds like: it represents an object within the network. In the context of recipes, it can be an ingredient or any object within a recipe. Furthermore, an object node is defined by two characters within a .txt file: O and S. O is the name of the object node whereas S is the state of the object. The name of the object which is preceded by O is followed by a 1 or 0 to denote whether the object is moving or not. The state can be anything such as a characteristic of the object, such as smooth, soft, hard, etc. It can also be where the object is currently located or what the object contains. For example, a bowl can be an object and its state could be that it contains some other objects and another one of its states could be that it is on top of some other object. So an object can have many states. For example:

O    carrot    0
S    orange
S    unpeeled

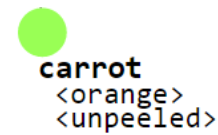

Fig 1. Sample object node of a carrot with two states. Images produced by FOON website visualizer [2].

As can be seen in the figure above, the sample object is a carrot with two states: orange and unpeeled. Object nodes are further divided into two types: input and output. Input and output object nodes are determined by a motion node, as discussed below. Object nodes with edges going into the motion node are input object nodes and any edges going out of the motion node go to output object nodes.

### B. Motion Node

Motion nodes are the other fundamental unit within FOON. They are the motions within the network and represent actions taken to transform the input object nodes into the output object nodes. Motions are defined by the M character within a .txt file and they are only defined by their name and a time stamp that denote the start time and end time of the action within a video. For example:

M    peel    <0:24,0:26>

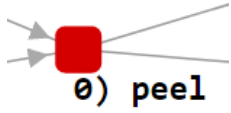

Fig 2. Motion node of peel with input and output edges.

As can be seen in the figure above, the motion node is peel; the number represents what order it was read in the file so 0) means it was the first motion read in the file. The edges with arrows coming in denote that input object nodes are inputs to the motion and the outgoing edges are the output object nodes of the motion.

*C. Functional Unit*

Combining the object nodes and the motion nodes together, the basic unit is formed called the functional unit [3]. A functional unit is composed of one or more input nodes that have edges going to a singular motion node, which outputs one or more output nodes. An image of a simple functional unit is shown:

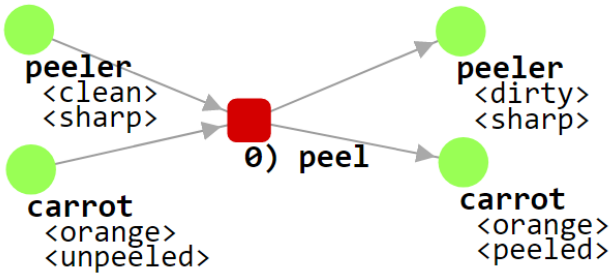

Fig. 3. Sample functional unit of peeling a carrot.

As can be seen, the functional unit above represents a simple action of peeling a carrot. Note that the state of the output has changed due to the action of peeling; the carrot has become peeled and the peeler has become dirty. In a .txt file, a function unit is represented between //, so for example, the .txt file version of the above function unit would be:

```
//
O    carrot    0
S    orange
S    unpeeled
O    peeler    1
S    clean
S    sharp
M    peel      <0:24,0:26>
O    carrot    0
S    orange
S    peeled
O    peeler    0
S    dirty
S    sharp
//
```

Hence, functional units are separated by //. Adding additional functional units for an entire recipe creates a subgraph.

*D. Subgraph*

A subgraph is a collection of functional units that total combine to make a single recipe. For example, Fig. 3. is a functional unit that may be in a subgraph for creating a vegetable stew. The unit of peeling a carrot is a single functional unit that when combined with many more result in a long chain of functional units that end with the final product of vegetable stew; this entire chain and combination of functional units makes up a subgraph and an entire .txt file represents this subgraph.

## III. VIDEO ANNOTATION AND FOON CREATION

*A. Manual Video Annotation*

In order to generate the entire FOON network, which is simply a collection of subgraphs, a human needs manually enter in data into .txt files to create subgraphs. In this regard, a person watches many recipe videos covering a range of foods and cuisine. While watching each video, the person records the steps taken within the video as functional units. At the end of the video, all the steps should have been converted into functional units and the entire listing of functional units made up the subgraph for that particular recipe.

*B. FOON Creation*

The manual video annotation was described above was done with many groups of people for 57 video recipes. The subgraphs of the video recipes were then combined into one large file that is the network. The FOON network is this combination of subgraphs across numerous videos. Code was run to ensure that there were no exact repeats of functional units as repeats do not provide any additional information.

*C. Supporting Data*

Supporting data was also utilized. Specifically, a .txt file containing all the motions within the FOON network and all of their success rates was provided. In practice, this file can be updated according to which actions the robot has little trouble to handle. The success rates were provided as probabilities between 0 and 1, so if this was implemented in practice, a user could increase the probability for actions that the robot would have no trouble doing and decrease the probability for actions that the robot would have difficulty doing (a human might be better for these motions). In total, there was 129 unique motions within the provided motion probability file, which is the same number of unique motions within the FOON network.

## IV. METHODOLOGY

After creating the main FOON network, the robot needs a way to traverse it such that it can reach all the steps it needs from a single goal node. Specifically, the algorithm that the robot utilizes to produce the ending recipe from FOON network is discussed here. Functions and base/class code provided (by Yu Sun and Md. Sadman Sakib) enabled data from the FOON network in a .txt file to be collected and organized into data structures including objects that represent functional units and its components as well as an object node class.

*A. General Search Algorithm*

The general search algorithm for traversing the FOON network generally follows this reasoning [4]. Firstly, the program takes in a goal node, which is where the search starts from. Secondly, it is placed into a queue in which the start of the queue is continually taken from until nothing can be taken from the queue (it is empty), which is when the algorithm ends. After an item is taken from the queue, it is then processed on. This includes mapping the object to potential functional units that contain that item as an output. Then, one functional unit is chosen from this list (heuristics discussed below determine which is chosen; in this case of iterative deepening search and the basic search algorithm, the first one is chosen). Then, the input nodes of that functional unit are placed into the queue unless there is an item which contains other ingredients (in which case, it is not ideal to count twice so only one time is taken). Additionally, they are also not placed in the queue if they are available in the kitchen, which is list of certain object nodes that are present already and so do not need to be searched further. All unique input objects are therefore placed in the queue and the functional unit is saved in a variable that records which functional units have been visited. Next, the loop starts over again until the queue is empty. At the end, when the loop terminates, the variable that recorded the functional units is reversed and printed into an output file, which is the steps of the recipe.

**Pseudocode for General Search Algorithm**

Starting with goal node G
Place G into queue structure Array
    Additions to array are made by adding to the back and items retrieved from Array are taken from the front of the structure (queue-like)
Until Array is empty, do the following:
   Take node out of Array from the front
   Check if node is available in kitchen, if not:
      Make list of candidate functional units that produce node as output
      ***Choose first candidate functional unit
      Add input nodes of candidate functional unit into Array
         Input object nodes should not be counted twice if one object contains other ingredients and other ingredients appear again as separate objects, so only unique input object nodes are added
      Append functional unit into recording variable Task Tree
      Repeat from beginning loop conditional
Reverse Task Tree
Output Task Tree contents into .txt file

*** - This instruction will change if utilization of Heuristic #1 or Heuristic #2 (see Section C and D below).

*B. Iterative Deepening Search*

The iterative deepening search utilizes the iterative deepening algorithm. For searching, the iterative deepening algorithm follows these steps. Firstly, there are variables that determine which depth a current node is on as well as a max node depth, which usually starts at 1. Next, the same steps as mentioned in the general search algorithm discussed in the prior section are taken, but if the search even takes out an object node that is an in a depth value greater than the max node depth, then the search is restarted, and the max node depth is incremented by 1. In this way, the depth is limited and the search will repeat until a max node depth is utilized in which the entire search is completed without going lower than that max node depth. Another difference is that a stack is utilized (since this is a version of depth-first search) instead so the input node in the queue is now a stack (so the next object is always taken from the back and added to the back now). Overall, this algorithm helps in that it provides a searching algorithm similar to a version of DPS but in which the depth is limited so that the program does not go down forever into a deep branch.

*C. Heuristic 1: Motion Success Rate*

The heuristic described here as well as in the next section is utilized in the section in the general search algorithm in which the program needs to make a choice as to which functional unit to choose to further search. In the iterative deepening search, the first functional unit is taken, but with a heuristic, it is possible to chose smartly as to which functional unit should be chosen and further searched with. The first heuristic implemented is based on the motion success rate [5]. Specifically, the program reads the success probabilities of each motion provided in a .txt file and then chooses the functional unit from the list that has the motion with the highest success rate. The rest of the search algorithm is the same as the basic search algorithm. This heuristic, however, helps in that it can provide a bias to choose the functional units that have actions that are more successful for the robot. Of course, the file with the motions and their probabilities can be tuned by a user such that this heuristic picks the motions with the most success rate for the robot to complete.

**Pseudocode for Choosing Candidate Functional Unit Based on Success Rate of Motion**

*** - Location in General Search Algorithm pseudocode
Set variable Max as -1
Set variable Index as -1
For each functional unit in candidate functional unit list:
    Check probability of motion success, and if it is more than Max, set Max to equal current probability and set Index variable as index of current functional unit
Continue General Search Algorithm steps with functional unit at the Index variable value (which is one with the most success probability)

*D. Heuristic 2: Least Input Ingredients*

The second heuristic implemented is the least input ingredients heuristic. This heuristic picks the optimal functional unit to search based on the least number of input objects. For example, if the list of functional units contains five functional units, this heuristic picks the one with the least number of input objects. The algorithm does this by going through the candidate functional units and counting the input ingredients. If the object contains other ingredients, then those ingredients are counted instead of the object. Additionally, if one ingredient appears within the contents are another and also is repeated as a separate object itself, then it is only counted once. In this way, an accurate count of how many input ingredients are needed are recorded so that the functional unit with the least number of input ingredients is chosen. Note that this might mean that the output step file might be larger, but this is because the algorithm disregards containers of other ingredients and repeats within the same functional unit so the total actual ingredients is still less but the text format of the functional unit may be larger.

### Pseudocode for Choosing Candidate Functional Unit Based on Least Input Object Nodes

\*\*\* - Location in General Search Algorithm pseudocode
Set variable Num = Infinity
Set Index as -1
For each functional unit in candidate functional unit list:
    Count number of input nodes – if there are container objects that contain other ingredients, count those ingredients and if so, do not count same ingredient if they appear as separate input object
    If count of input nodes < Num:
        Set Num as count of input nodes
        Set Index as index of current functional unit
Continue General Search Algorithm steps with functional unit at the Index variable value (which is one with the least number of input object nodes)

## V. EXPERIMENT & DISCUSSION

After experimentation with 5 goal nodes, some patterns and discoveries emerged. Overall, there was no problem as long as the goal node went through the FOON network and eventually reached object nodes that were available in the kitchen (which would make the recipe possible).

### A. Number of Functional Units in Task Tree

The number of functional units within the final outputted task tree varied across the three algorithms. If counting the general algorithm, it does its job in that the functional units could be followed to produce the outputted final goal node object. In terms of the number of functional units for the unique algorithms, the IDS (iterative deepening search) algorithm and the general search algorithm sometimes produce the exact same number of functional units in the output. Overall, the IDS algorithm helps to prevent going down a deep branch so the final output is at time identical to the general algorithm, which is a BFS algorithm.

The heuristic algorithms aid in choosing the candidate functional unit from a list of potential ones, so if there is always only one potential functional unit, the number of functional units will be the same across both the algorithms. However, for Heuristic 1, since it depends on the success rate of the motion, the number of functional units within the output really depends on the success rate. If one functional unit has a motion with a higher success rate but many, many unique object nodes, then it will have a larger number of functional units in its output than a functional unit with a motion with a lower success rate and a lower number of input object nodes.

In that same sense, Heuristic 2 aims to choose the functional unit with the least number of input object nodes. However, it only counts the unique input object nodes. So choosing a functional unit with the lowest unique input object nodes does not mean that it is the shortest in the number of lines it takes up, as the heuristic does not count repeats but the entire functional unit (including repeats) still appears in the final output. Additionally, choosing a unit with the fewest input nodes might mean that those nodes are simply more complex ingredients which could branch out to many, many more simpler input nodes. Since we are only seeing down one level and choosing the optimal one (least unique input nodes), we do not know if two or three levels down if it might spread out to a greater number of simpler input objects.

### B. Time and Space Complexity

b refers to the branching factor at each non-leaf node and d is defined as the depth that the solution is found at. For the general search algorithm, it is based on coded with a BFS mindset so that time complex for that is $O(b^d)$ and the space complexity is $O(b^d)$.

IDS has a time complexity of $O(b^d)$ and a space complexity of $O(b*d)$. The difference between IDS and a regular BFS is that the space complexity goes from $O(b^d)$ for a BFS to $O(bd)$ for IDS, so it goes down. Overall, the IDS algorithm has the advantages of the BFS algorithm in that it has completeness (something a regular DFS does not have) but it also has the better space complexity associated with a DFS (of $O(bd)$), which is the best of both worlds.

For the heuristic functions, since they are simply utilizing the general search algorithm structure but only optimizing for which success rate is highest as well as lowest unique input nodes, it sort of acts like there are weights associated with each edge. This results in a time complexity of $O(b^d)$ and a space complexity of $O(b^d)$ as for a regular BDS and since a BDS has completeness, it is good choice of base structure for the heuristic programs.

## VI. CONCLUSION & FUTURE WORK

These algorithms as well as the construction of the FOON network displays a potential structure that robots can utilize to make decisions of their own regarding cooking and recipe generation. Of course, the structure of FOON and its implementations can range beyond what this paper dives into; specifically, the aspect of input, motion, and output does not have to be limited to simple cooking actions but can generally be extended to other possibilities that following the same cause and effect relationship. Such new alteration to the FOON for new situations can provide different ways that robots can be utilized in other aspect of life to aid humans.